\documentclass{article}
\usepackage{spconf,amsmath,graphicx}
\usepackage{cite}
\usepackage{amssymb,amsfonts}
\usepackage{algorithmic}
\usepackage{textcomp}
\usepackage{xcolor}
\usepackage{multirow}
\usepackage{booktabs}
\usepackage{soul}
\usepackage{subcaption}
\usepackage{url}

\title{UNCERTAINTY-AWARE AB3DMOT BY VARIATIONAL 3D OBJECT DETECTION}
\name{Illia Oleksiienko and Alexandros Iosifidis}
\address{DIGIT, Department of Electrical and Computer Engineering, Aarhus University, Denmark \\
\{io,ai\}@ece.au.dk}
\begin{document}
%
\maketitle
\begin{abstract}
Autonomous driving needs to rely on high-quality 3D object detection to ensure safe navigation in the world. Uncertainty estimation is an effective tool to provide statistically accurate predictions, while the associated detection uncertainty can be used to implement a more safe navigation protocol or include the user in the loop. In this paper, we propose a Variational Neural Network-based TANet 3D object detector to generate 3D object detections with uncertainty and introduce these detections to an uncertainty-aware AB3DMOT tracker. This is done by applying a linear transformation to the estimated uncertainty matrix, which is subsequently used as a measurement noise for the adopted Kalman filter. We implement two ways to estimate output uncertainty, i.e., internally, by computing the variance of the CNN outputs and then propagating the uncertainty through the post-processing, and externally, by associating the final predictions of different samples and computing the covariance of each predicted box. In experiments, we show that the external uncertainty estimation leads to better results, outperforming both internal uncertainty estimation and classical tracking approaches. Furthermore, we propose a method to initialize the Variational 3D object detector with a pretrained TANet model, which leads to the best performing models.
\end{abstract}

\section{Introduction}
3D Object Detection (3D OD) is the problem that aims to detect objects in the 3D world, providing the coordinates relative to the sensor and true world sizes of the detected objects. In contrast to 2D Object Detection which is commonly based on the appearance of objects in an image or video frame, in 3D OD the size of objects is not distorted by projections and rigid objects retain their size on every data frame.
However, to obtain the data that can accurately represent objects in 3D requires specialized sensors. 
Even though color images \cite{2016mono3dad, 2019mono3dpp, 2018mf3d, 2018monogrnet, 2018oftnet} or stereo images \cite{2018pseudolidar, 2019pseudolidar, 2020cdn} can be utilized to obtain 2D-like object detections and use those to estimate the corresponding 3D bounding boxes, the detection accuracy and speed of such methods is inferior to that of methods utilizing point cloud data captured by Lidar sensors, such as VoxelNet \cite{2017voxelnet}, SECOND \cite{2018second}, PointPillars \cite{lang2018pointpillars} and TANet \cite{liu2019tanet}.
\begin{figure}
     \centering
     \begin{subfigure}[b]{1\linewidth}
         \centering
         \includegraphics[width=0.9\linewidth]{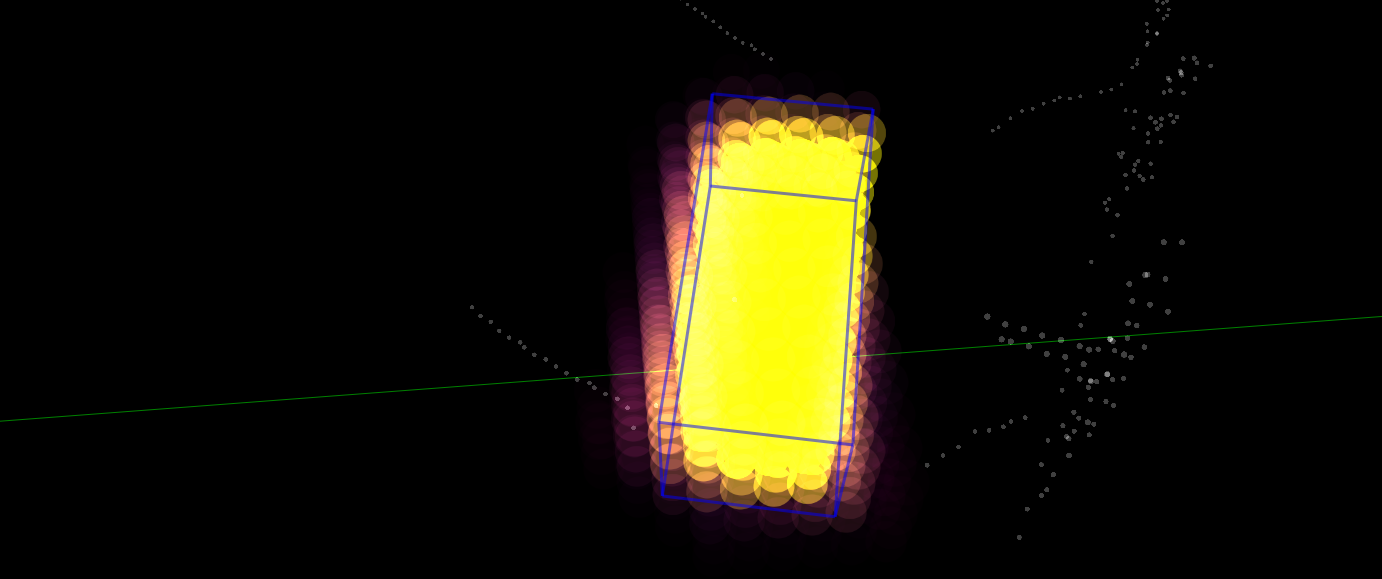}
         \caption{3D bounding box with the uncertainty provided by the Variational TANet 3D object detector}
     \end{subfigure}
     \begin{subfigure}[b]{1\linewidth}
         \centering
         \includegraphics[width=0.9\linewidth]{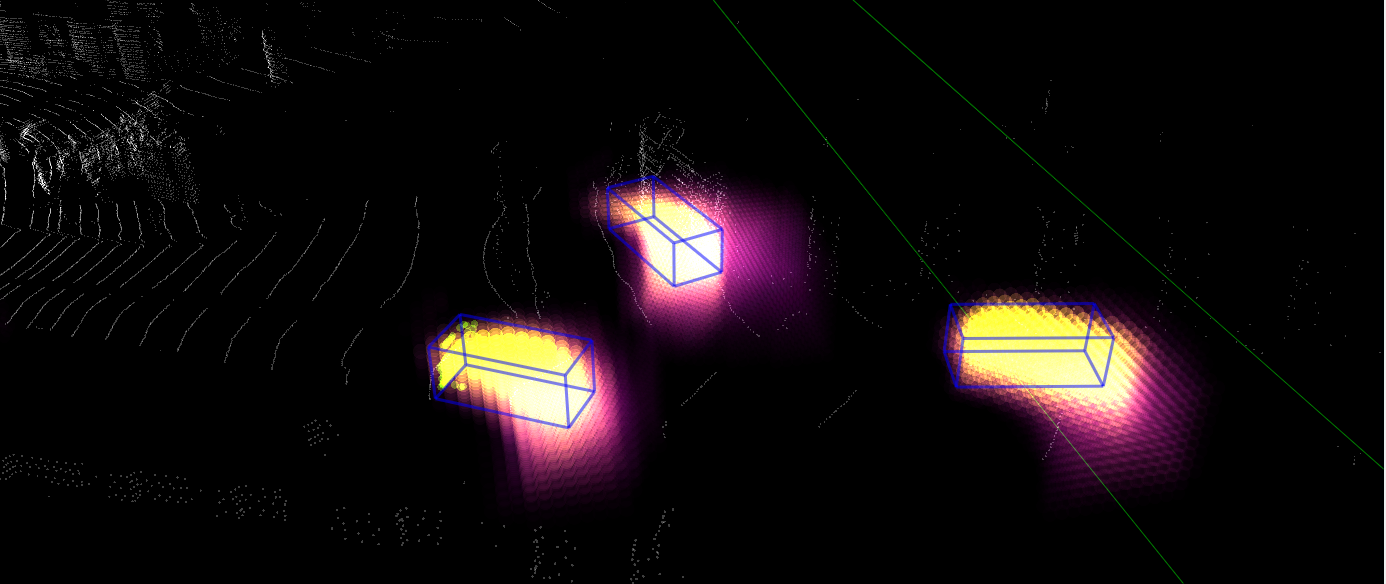}
         \caption{3D bounding boxes with uncertainties provided to the Kalman filter of the uncertainty-aware AB3DMOT tracker}
     \end{subfigure}
        \caption{Examples of 3D bounding boxes with raw uncertainties obtained by the proposed Variational TANet (top figure), and the uncertainties provided to the Kalman filter of the proposed Uncertainty-aware AB3DMOT (bottom figure).
        Yellow color represents higher probability, while the purple color represents lower probability. Blue boxes correspond to the mean predictions.}
        \label{fig:uncertainties}
\end{figure}

3D Multiple Object (3D MOT) tracking aims to not only find the position of objects in the 3D world, but also assign a unique ID to each of them for the entire duration that an object appears in the data sequence. 
3D MOT can be performed in a single stage, where the object detection in each frame and objects associations in successive frames are performed by the same model \cite{2020centerbased3dod, 2018jointmono3d, 2019complexeryolo}, or in two stages, where the detections are generated by a 3D OD model and the object associations in successive frames are performed separately.

Most 3D OD methods exploit object detections provided by powerful object detectors, commonly based on deep neural networks providing point estimates of their outputs. While recent advances in deep learning have led to remarkable results, this approach strongly restricts the ability of the object association step to exploit possible uncertainties in the object detections. Moreover, there has been some evidence that the predictive uncertainty of accurate deep learning models does not necessarily correlate with their confidence in their outputs \cite{guo2017calibration, minderer2021calibration}. 
Uncertainty estimation in neural networks allows using the uncertainty in predictions to perform better decision-making, which is important in critical fields such as medical image analysis or autonomous driving.
The practical application of uncertainty estimation includes 3D Object Tracking \cite{zhong2020uavoxel, wang2020ua3dself}, 3D Object Detection \cite{feng20183ddropout, meyer2019lazernet, meyer2020ua3ddetlabels}, 3D Human Pose Tracking \cite{daubney2011ua3dhuman}, Steering Angle Prediction \cite{loquercio2020uncertaintydriving}, providing better prediction and control than without the use of uncertainty.

In this paper, we propose an uncertainty-aware 3D object tracking pipeline. We show how the TANet 3D object detector \cite{liu2019tanet} can be formulated and trained as a Variational Neural Network \cite{oleksiienko2022vnn} to generate object detections with uncertainty in two ways, i.e., internally with variance estimation, and externally with covariance estimation. We combine them with a two-stage 3D MOT method called AB3DMOT \cite{2020ab3dmot} that utilizes a 3D Kalman filter \cite{1960kalmanfilter} for state prediction and Hungarian algorithm \cite{1955hungarian} for objects' association. We modify AB3DMOT to use the predicted output uncertainty of the 3D OD model in the 3D Kalman Filter and show that this improves the Multi Object Tracking Accuracy (MOTA), F1 score and Mostly Lost (ML) metrics.

The remainder of the paper is structured as follows. Section \ref{S:RelatedWork} describes related and prior work. In Section \ref{S:ProposedMethod} we describe the proposed approach, including the Variational TANet model and its training and the proposed uncertainty-aware AB3DMOt. Section \ref{S:Experiments} outlines the experimental protocol and provides experimental results. Section \ref{S:Conclusions} concludes this paper\footnote{Our code is available at \url{gitlab.au.dk/maleci/opendr/ua-ab3dmot}}.

\begin{figure*}[!ht]
\centering
    \includegraphics[width=\linewidth]{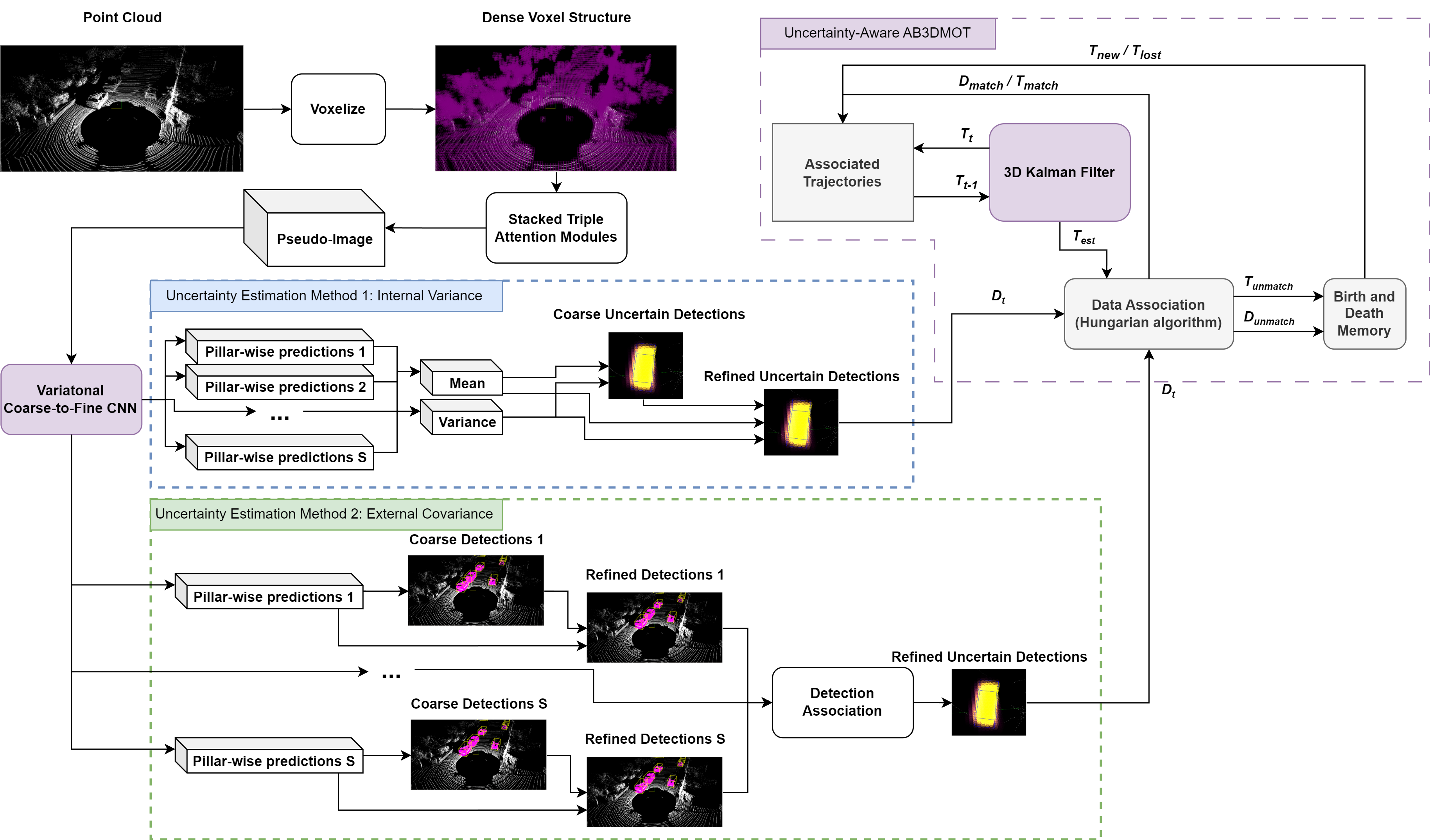}
    \caption{The pipeline of the proposed Uncertainty-Aware AB3DMOT by Variational 3D Object Detection method. Following TANet \cite{liu2019tanet}, the input point cloud is voxelized to create a pillar map and processed by a Stacked Triple Attention Module to generate a 2D pseudo-image. This pseudo image is used by a Variational version of the Coarse-to-Fine convolutional network to create pillar-wise predictions. Uncertainty from these predictions is estimated by combining outputs of multiple applications of this network to the same inputs in one of two ways. The Internal Variance approach computes variance of pillar-wise predictions and propagates it through the post-processing steps, while the External Covariance approach computes the final prediction for each sample and then applies an association algorithm to find groups of object detections corresponding to the same object to compute the covariance between them. Uncertain object detections are passed to an Uncertainty-Aware AB3DMOT, which applies a linear transformation to the uncertainty and uses it as an additional value to a Kalman filter. A Kalman filter is used to predict changes in object positions between frames, and Hungarian algorithm is used to associate new object detections with known tracklets.}
    \label{fig:uaab3dmot}
\end{figure*}

\section{Related Work}\label{S:RelatedWork}
3D OD based on Lidar point clouds cannot be performed with standard CNNs, as the point cloud data consists of a set of irregular 3D point positions. In order to introduce point cloud data as inputs to Deep Learning methods, one needs to structure it first, with voxelization being the most popular way of doing so.
Voxelization is performed by selecting a region of interest in the 3D space and splitting it using a grid of same-sized cuboid shapes, called voxels.
VoxelNet \cite{zhong2020uavoxel} creates a 3D grid of voxels and processes points inside each voxel to create voxel features, which are later processed by 3D and 2D convolutional layers. 
To accelerate the process, PointPillars \cite{lang2018pointpillars} uses voxels with the maximum vertical size called pillars, resulting in a 2D grid that can be used as input to a 2D CNN model, which is much faster than using 3D convolutions.
TANet \cite{liu2019tanet} introduces the Triple Attention module for extracting better features from pillars and a Coarse-to-Fine regression module that produces coarse detections first and then refines them with an additional subnetwork.

The fastest 2D Multiple Object Tracking (MOT) method called SORT \cite{bewley2016sort} utilizes 2D Kalman Filter \cite{1960kalmanfilter} and Hungarian algorithm \cite{1955hungarian} to associate the 2D object detections provided by a Deep Learning model. The 3D version of SORT, called AB3DMOT \cite{2020ab3dmot}, performs better than SORT in 3D MOT by utilizing a 3D Kalman Filter instead of a 2D one. 
In \cite{zhong2020uavoxel}, the use of uncertainty for 3D MOT is investigated by training the voxel-based 3D OD method SECOND \cite{2018second} with an additional branch to predict the model's uncertainty, Then, the predicted uncertainty is used in the Kalman filter's measurement noise leading to better tracking results. The use of predicted uncertainty, instead of assuming a unit-Gaussian distribution over the measurements expressed by using the identity matrix, for measurement noise is also supported by \cite{russell2022multivariateuncertaintykalman}.

Uncertainty estimation in neural networks can be performed using different ways to model uncertainty.
The four main categories of uncertainty estimation methods defined in \cite{gawlikowski2021uncertaintyindl} are: Deterministic Methods \cite{2018evidentialdl, zhong2020uavoxel} which use a single network pass with uncertainty estimated from one of the network branches or by analyzing the behavior of the network; Bayesian Neural Networks (BNNs) \cite{blundell2015weight, magris2023BNNsurvey} that sample different models from the corresponding weight distribution and process the input using them resulting in a set of predictions, mean and variance of which are used to estimating the predictive distribution; Ensemble Methods \cite{osband2018randomized, valdenegro2019subens}, that can also be seen as BNNs with Categorical distribution; and Test-Time Data Augmentation methods \cite{wang2018barintumortesttimeaug, wang2019aleamedical, kandel2021testtime} that apply different augmentations to the input to create a set of slightly different inputs that are processed by the same model.

While Deterministic Methods require a single pass of the network leading to lower inference time, they usually perform worse than the rest of the methods as they can be seen as a point estimation of statistically better Bayesian approaches. 
Variational Neural Networks \cite{oleksiienko2022vnn,oleksiienko2022vnntorchjax} are similar to BNNs, but consider a Gaussian distribution over the outputs of each layer, the mean and variance of which are generated by the corresponding sub-layers. 
The output uncertainty is calculated by sampling different values from the corresponding Gaussian distributions, resulting in multiple predictions for the same input.

\section{Uncertainty-aware 3D Object Tracking}\label{S:ProposedMethod}

\subsection{Variational TANet}
TANet \cite{liu2019tanet} is a pillar-based 3D Object Detection model that improves PointPillars by introducing the Triple Attention module to extract better pillar (voxel) features and by using the Coarse-to-Fine fully convolutional network to process the voxel pseudo image and generate the bounding box predictions. 
TANet is an anchor-based method \cite{shaoqing2015fasterrcnn}, which means that the output from the Coarse-to-Fine network is a voxel-wise prediction with position and sizes predicted as offsets from the corresponding anchor.
This output is then processed by a Non-maximum-suppression module to select the most likely boxes and decoded based on the anchors to create the final 3D bounding boxes. 

We create a Variational TANet (VTANet) model by replacing the fully convolutional Coarse-to-Fine module with a Variational Neural Network (VNN) \cite{oleksiienko2022vnn, oleksiienko2022vnntorchjax} based version of it.
The training is performed using the same loss function and training procedure, except we use multiple model samples per each data input and train for the mean of the predictions of all samples. 
The structure of the proposed method is shown in Fig. \ref{fig:uaab3dmot}.
Variational Neural Networks are selected as the choice of uncertainty estimation method because they have shown to achieve high uncertainty quality, low memory consumption and ease of application to an existing model architecture \cite{oleksiienko2022vnn}.

We propose two ways for the VTANet model to provide its estimated uncertainty to the subsequent tracker.
The first, namely internal, way computes mean and variance of predictions of the Variational Coarse-to-Fine network and provides the variance through the decoding stage by treating the values as Gaussian random variables. 
The outputs of the CNN are voxel-wise in the form of $(x, y, z, w, l, h, r)$ with respect to an anchor of the same data format, where $(x, y, z)$ is the relative position vector, $(w, l, h)$ is the relative size vector and $r$ is the relative rotation angle.
For a set of predictions from $(x_i, y_i, z_i, w_i, l_i, h_i, r_i), i \in [1, S]$, where $S$ is the number of samples, we compute the mean prediction $(m_x, m_y, m_z, m_w, m_l, m_h, m_r)$ and the corresponding variance $(v_x, v_y, v_z, v_w, v_l, v_h, v_r)$. The anchor $(x_a, y_a, z_a, w_a, l_a, h_a, r_a)$ that corresponds to the prediction is used to create the decoded predictions $(\hat{x}, \hat{y}, \hat{z}, \hat{w}, \hat{l}, \hat{h}, \hat{r})$ as follows:
\begin{align}
\begin{split}
    & d = \sqrt{l_a^2 + w_a^2}, \\
    & (\hat{x}, \hat{y}) = (m_x, m_y) d + (x_a, y_a), \\
    & (\hat{w}, \hat{l}, \hat{h}) = (e^{m_w} w_a, e^{m_l} l_a, e^{m_h} h_a), \\
    & \hat{r} = m_r + r_a, \\
    & \hat{z} = m_z h_a + z_a + h_a / 2 - \hat{h} / 2.\\
\end{split}
\end{align}
The corresponding variances $(v_{\hat{x}}, v_{\hat{y}}, v_{\hat{z}}, v_{\hat{w}}, v_{\hat{l}}, v_{\hat{h}}, v_{\hat{r}})$ are calculated as follows:
\begin{align}
\begin{split}
    & d = \sqrt{l_a^2 + w_a^2}, \\
    & \mathrm{v_{exp}}(m, v) = e^{2m + 2v} - e^{2m + v}, \\
    & (v_{\hat{x}}, v_{\hat{y}}) = (v_x, v_y) d^2, \\
    & (v_{\hat{w}}, v_{\hat{l}}, v_{\hat{h}}) = (\mathrm{v_{exp}}(\hat{w}, v_w) w_a^2, \mathrm{v_{exp}}(\hat{l}, v_l) l_a^2, \mathrm{v_{exp}}(\hat{h}, v_h) h_a^2)\\
    & \hat{r} = v_r, \\
    & v_{\hat{z}} = v_z h_a^2,
\end{split}
\end{align}
where $\mathrm{v_{exp}}(m, v)$ is the variance of the exponent of the Gaussian random variable.

The second, namely external, way to compute uncertainty runs the full pipeline of TANet to create $S$ 3D bounding box predictions, where $S$ is the number of neural network parameter samples. 
These predictions are then grouped by finding the best association for each predicted object, similarly to the Hungarian algorithm on the tracking step. 
Consider the set of predictions $P^S=\{p_i \:|\: i \in [1, \dots, S]\}$ where each element $p_i$ is a set of 3D bounding boxes $p_i=\{b_k \:|\: k \in [1, \dots, K_i] \}$, with $K_i$ being the number of predicted boxes for the sample $i$.
For different neural network parameter samples, some boxes with high uncertainty may or may not appear, which results in slightly different values of $K_i$.
We define the association set $A(P)$ as a set of box groups $\{g_q\}$ where each group consists of 3D bounding boxes from different samples with the closest distances to each other, but no more than $1$ meter.
The groups are created by iterating through all boxes for each sample and checking the smallest distance to the average position of each existing group. If the distance is lower than $1$ meter, the object is added to the selected group, and otherwise it creates a new group:
\begin{align}
\begin{split}
    &\forall i \in [1 .. S] \: \forall k \in [1 .. K_i]\\
    &\hat{G}(b_k) = \underset{g_q}{\mathrm{argmin}} \:\: |\mathrm{avg}(\{b^{\mathrm{x,y,z}} \in g_q\}) - b^{\mathrm{x,y,z}}_k|\\
    &G(b_k) =
    \begin{cases}
        \hat{G}(b_k), & \text{if } |\mathrm{avg}(\{b^{\mathrm{x,y,z}} \in \hat{G}(b_k)\}) - b^{\mathrm{x,y,z}}_k| \leq 1, \\
        g_{\text{new}}, & \text{otherwise},
    \end{cases} \\
    & g_q = \{b_k \:|\: G(b_k) = q, \: i \in [1 .. S], \: k \in [1 .. K_i]\},
\end{split}
\end{align}
where $b^{x,y,z}$ is a position of the bounding box $b$ and $\mathrm{avg}(\cdot)$ is the averaging function.
For each group $g_q$, we compute the mean bounding box and the covariance matrix in the form of a $7 \times 7$ matrix for position $(x, y, z)$, size $(w, h, l)$ and rotation $\alpha$.

\subsection{Initialization of Variational TANet by pretrained model}
Instead of training the VTANet model from scratch, we can initialize it based on an already trained TANet model, which provides point estimates.
This can be done based on the fact that infinitely small variance in a Gaussian distribution transforms it into a Dirac delta distribution, with all distributional mass placed on the mean value.
By following the reverse process, we initialize the VTANet model with means from the corresponding pretrained TANet parameters, and the variance weights are set to be small values by either filling them with constant values or by using Xavier normal or uniform initialization \cite{glorot2010xavierinit}.
Small variance values will not deviate too much from the initial pretrained values, but allow for training the whole model together and to find the optimal variance and mean parameters.
We refer to a VTANet model trained by means of transfer learning from a pretrained TANet as IVTANet hereafter.
\begin{table*}[!ht]
    \caption{Tracking results on KITTI tracking dataset.}
    \label{tab:tracking-results}
    \centering
    \begin{tabular}{l|p{1.7cm}p{1.5cm}p{1.5cm}p{2.5cm}|cccc}
        \toprule
        \textbf{Model} & \textbf{Uncertainty method} & \textbf{Training Samples} & \textbf{Inference Samples} & \textbf{Tracking \break Parameters} & \textbf{MOTA\%$\uparrow$} & \textbf{F1\%$\uparrow$} & \textbf{ML\%$\downarrow$}\\
        \midrule
            Voxel von-Mises \cite{zhong2020uavoxel} & deterministic & - & - & SORT \cite{bewley2016sort} & - & 55.10 & 30.60 \\ \hline
            TANet \cite{liu2019tanet}& - & - & - & AB3DMOT \cite{2020ab3dmot} & 68.71 & 85.69 & 8.58 \\
            IVTANet + UA-AB3DMOT & covar & 2 & 4 & $\alpha=0.6, \: \beta=5$ & \textbf{72.30} & \textbf{87.34} & \textbf{7.74}\\
            IVTANet + UA-AB3DMOT & covar & 2 & 4 & $\alpha=0, \: \beta=1$ & 72.13 & 87.26 & 7.74\\
            IVTANet + UA-AB3DMOT & covar & 2 & 4 & $\alpha=1, \: \beta=0$ & 72.05 & 87.22 & 7.74\\
            VTANet + UA-AB3DMOT & covar & 3 & 3 & $\alpha=0.6, \: \beta=5$ & 69.63 & 86.46 & 7.95\\
            VTANet + UA-AB3DMOT & covar & 3 & 3 & $\alpha=0, \: \beta=1$ & 69.48 & 86.39 & 8.16\\
            VTANet + UA-AB3DMOT & covar & 3 & 4 & $\alpha=0.6, \: \beta=5$ & 69.42 & 86.38 & 9.00\\
            VTANet + UA-AB3DMOT & covar & 3 & 3 & $\alpha=1, \: \beta=0$ & 69.15 & 86.31 & 9.00\\
            VTANet + UA-AB3DMOT & covar & 3 & 4 & $\alpha=1, \: \beta=0$ & 68.86 & 85.04 & 9.21\\
            VTANet + UA-AB3DMOT & covar & 2 & 4 & $\alpha=0.6, \: \beta=5$ & 68.54 & 86.13 & 8.16\\
            VTANet + UA-AB3DMOT & var & 3 & 4 & $\alpha=0.5, \: \beta=5$ & 68.46 & 85.91 & 7.95\\
        \bottomrule
    \end{tabular}
\end{table*}

\subsection{Uncertainty-Aware AB3DMOT}

As shown in \cite{zhong2020uavoxel, russell2022multivariateuncertaintykalman}, the Kalman filter benefits from providing actual uncertainties instead of assuming a unit-Gaussian distribution over the measurements.
We follow this approach and provide the modified variance diagonal matrices or covariance matrices from the predicted uncertainties to the Kalman filter.
We modify the uncertainty matrices provided by the detector with a linear combination:
\begin{equation}
    \hat{\Sigma} = \alpha I + \beta \Sigma,
\end{equation}
where $\hat{\Sigma}$ is the uncertainty provided to the Kalman filter, $\Sigma$ is the uncertainty predicted by the VTANet object detector, $\alpha$ is a hyperparameter used to control the contribution of the base uncertainty to $\hat{\Sigma}$, and $\beta$ is a hyperparameter that controls the contribution of the predicted uncertainty to $\hat{\Sigma}$.
This transformation aims to provide a degree of freedom for the Kalman filter in using the predicted object detections while utilizing the actual uncertainties in the predictions. Using the values $\alpha = 1$ and $\beta = 0$ leads to the standard AB3DMOT exploiting unit-Gaussian distribution measurement noise in the Kalman filter. 
Fig. \ref{fig:uncertainties} illustrates examples of uncertainties estimated from the detector and the modified uncertainty provided to the Kalman filter.

\section{Experiments}\label{S:Experiments}
We train TANet and VTANet models using the standard training procedure on KITTI \cite{2012kitti} dataset described in \cite{liu2019tanet}. 
VTANet models are trained with sample count in $[1, \dots, 4]$ range and each model is evaluated with AB3DMOT on KITTI tracking dataset with every number of samples from the same range.
We train the internal uncertainty VTANet models that provide the variance values and the external uncertainty models that provide the covariance values.
Each model is evaluated with the different configuration of uncertainty transformation parameters $\alpha$ and $\beta$, including $\alpha=0$, $\beta=1$ for the predicted uncertainty only, $\alpha=1$, $\beta=0$ to not use the predicted uncertainty and different combinations of $\alpha \in [0, 1]$ and $\beta \in \{0.1, 1, 5, 10, 50\}$.
Additionally, we train IVTANet models which are initialized with a pretrained TANet values for means and small variance weights and trained using the same training procedure as VTANet models. We compare the performance of the proposed method with that of the Voxel von-Mises method \cite{zhong2020uavoxel}, which is the only method we found in the literature incorporating uncertainty estimation in 3D object tracking. This method employs SECOND \cite{2018second} for 3D OD and SORT \cite{bewley2016sort} for objects associations.

Table \ref{tab:tracking-results} shows the performance of the competing models on the KITTI tracking dataset based on three performance metrics, namely the Multi Object Tracking Accuracy (MOTA), F1 score, and Mostly Lost (ML). This table also provides information on the hyperparameter values used for different model, i.e., the type of adopted uncertainty estimation, the number of network parameter samples used during training and inference, and the tracking method and hyperparameter values. 

As can be seen, the proposed uncertainty-aware 3D MOT method with a VTANet trained from scratch, i.e., VTANet + UA-AB3DMOT, provides better MOTA, F1 and ML scores compared to the classic 3D MOT approach combining TANet with AB3DMOT, i.e., TANet + AB3DMOT, due to the use of uncertainty. 
The external way to compute uncertainty as covariance provides better results compared to the internal one providing variance across all configurations, and therefore, only a sub-set of configurations is provided in the results. The Voxel von-Mises method \cite{zhong2020uavoxel} provides much worse results due to the use of inferior methods for each of its processing steps, i.e., it adopts the less accurate 3D detector SECOND \cite{2018second}, the less suited for 3D object detection tracker SORT \cite{bewley2016sort} and a deterministic uncertainty estimation method that does not rely on a BNN or an Ensemble method during training.
The best MOTA, F1 and ML scores are achieved by the configurations of the proposed method which employ a VTANet pretrained using a TANet providing point estimates. Those are denoted by IVTANet + UA-AB3DMOT in Table \ref{tab:tracking-results}. As can be seen in the results, the use of an IVTANet 3D object detector improves performance compared to the classic 3D MOT approach even without exploiting the estimated uncertainty (i.e., when $\alpha=1$ and $\beta=0$). Overall, the highest performance is achieved by using a VTANet obtained by training a pretrained TANet combined with the AB3DMOT model exploiting the estimated uncertainty (IVTANet + UA-AB3DMOT). It is interesting to see that the use of the same linear combination values ($\alpha = 0.6$ and $\beta = 5$) leads to the best performance when both VTANet and IVTANet 3D object detectors are used.

\section{Conclusions}\label{S:Conclusions}
In this paper, we proposed an uncertainty-aware 3D object tracking pipeline that utilizes a Variational Neural Network-based version of TANet 3D object detector to generate predictions with uncertainty in two different ways, i.e., internally with variance estimation and externally with covariance estimation. 
Predictions from the detector are introduced into the uncertainty-aware tracker that utilizes the estimated uncertainty to determine the measurement noise in the Kalman filter, leading to an improvement in MOTA, F1 and ML metrics.
We also proposed an effective way to perform transfer learning from pretrained TANet to the proposed VTANet. This way, effective existing models can be used to create corresponding variational models.
The resulting model is trained using the regular training procedure and leads to a much better tracking performance compared to the models trained from scratch.

\begin{small}
\section*{Acknowledgement}
This work has received funding from the European Union’s Horizon 2020 research and innovation programme under grant agreement No 871449 (OpenDR). This publication reflects the authors’ views only. The European Commission is not responsible for any use that may be made of the information it contains.
\end{small}

\bibliographystyle{IEEEbib}
\bibliography{bibliography.bib}

\end{document}